\documentclass[10pt,twocolumn,letterpaper]{article}

\usepackage{cvpr}
\usepackage{times}
\usepackage{epsfig}
\usepackage{graphicx}
\usepackage{amsmath}
\usepackage{amssymb}

\usepackage[pagebackref=true,breaklinks=true,letterpaper=true,colorlinks,bookmarks=false]{hyperref}

\cvprfinalcopy 

\begin{document}

\title{Bounding Box Embedding for Single Shot Person Instance Segmentation}

\author{Jacob Richeimer \qquad Jonathan Mitchell
\\
Octi Inc.\\
{\tt\small \{jacob, jonathan\}@octi.tv}
}

\maketitle

\begin{abstract}
   We present a bottom-up approach for the task of object instance segmentation using
   a single-shot model. The proposed model employs a fully convolutional
   network which is trained to predict class-wise segmentation masks as well
   as the bounding boxes of the object instances to which each pixel belongs.
   This allows us to group object pixels into individual instances.
   Our network architecture is based on the DeepLabv3+ model ~\cite{Deeplabv3+},
   and requires only minimal extra computation to achieve pixel-wise instance assignments.
   We apply our method to the task of person instance segmentation, a common task relevant
   to many applications.   
   We train our model with COCO data~\cite{COCO} and report competitive results for the
   person class in the COCO instance segmentation task.
\end{abstract}

\newcommand{\bfxi}{\ensuremath{\mathbf{x}_i}}

\section{Introduction}

Huge advances have recently been made in the application of convolutional networks to the localization of objects in images, allowing real-time deployment of such models even with restrictive hardware limitations such as on mobile devices. In applications which call for a high level of scene understanding, such as robotics and augmented reality, models which can be designed for efficiency and employed at low computational cost are crucial.

The development of single shot approaches to object detection, such as SSD, YOLO, and RetinaNet~\cite{SSD, YOLO9000, Retinanet} allow for objects to be localized up to their bounding boxes within a given image. The FCN, U-Net, and DeepLab architectures~\cite{FCN, Unet, Deeplabv3, Deeplabv3+}, among others, developed for semantic segmentation, allow for pixel-wise categorical classification.

\begin{figure}
	\begin{center}
		\includegraphics[width=0.75\linewidth]{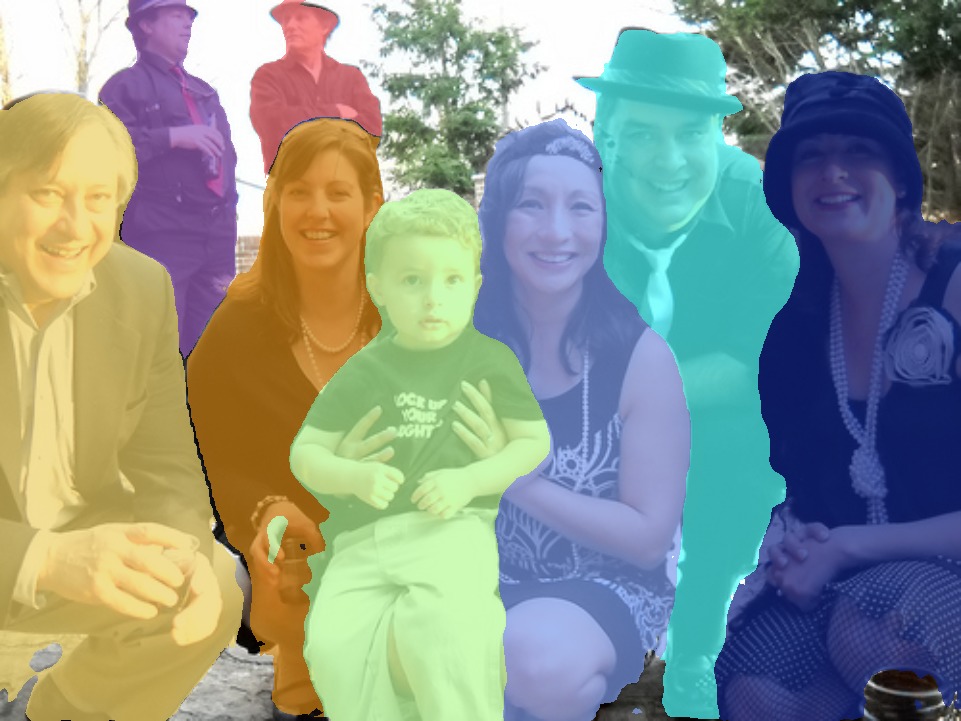}
	\end{center}
	\caption{Visualization of the results of our method on a sample image with a crowded scene of person instances.}
	\label{fig:anchorbox}
\end{figure}

\begin{figure*}
	\begin{center}
		\includegraphics[width=0.9\textwidth]{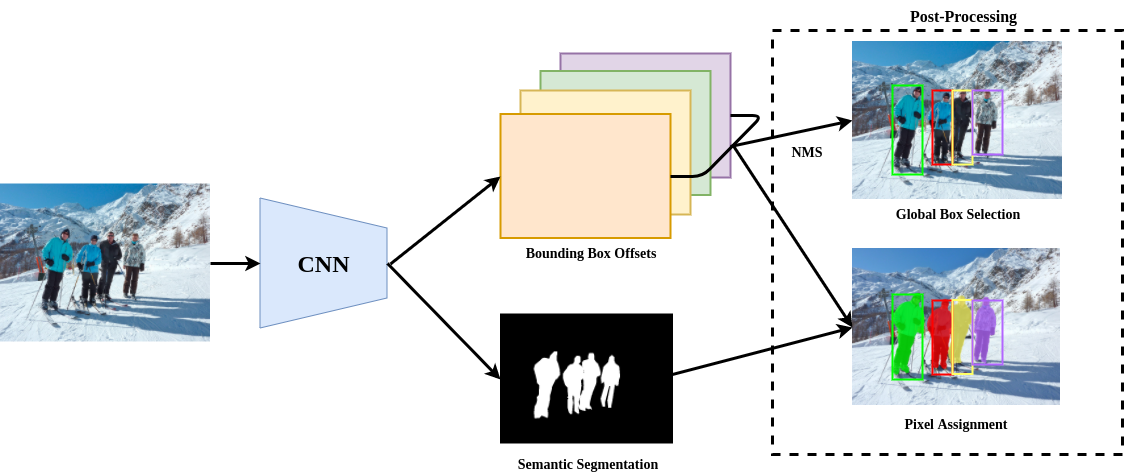}
	\end{center}
	\caption{
		Our Bounding Box Embedding method consists of an off-the-shelf CNN model which is trained to predict (1) A semantic segmentation mask, and (2) per-pixel instance bounding box offsets from prior ``anchor" boxes. The output boxes are used to predict bounding boxes for object instances in the image, similar to many existing single-shot object detection systems. The segmentation mask is then used together with the pixel-wise bounding box predictions to assign each pixel to a detected object instance.	
	}
	\label{fig:bbe-diagram}
\end{figure*}

However, the task of instance segmentation is often necessary for a deeper understanding of visual scenes, and is more fine-grained than both of the above. It demands for each pixel, not only a class label, but also the identification of to which object instance of that class it belongs.

The solutions to instance segmentation, similar to those for multi-person pose estimation, can be conceptually divided between ``top-down" and ``bottom-up" approaches. Top-down approaches focus first on locating object instances, followed by obtaining pixel-wise masks for each detected instance. Bottom-up approaches, on the other hand, first determine the object class of each pixel, then afterwards focus on grouping the pixles into individual instances.

The current state-of-the art approaches to the task of instance segmentation are still dominated by the ``top-down" approach, which relies on a prior object detection step or on a region proposal sub-network to be built in to the architecture. This can be computationally heavy and often quite difficult to employ in real-time applications.

Lately, the computer vision community has been focusing more effort on the instance segmentation task, and more ``bottom-up" approaches have been proposed. In this paper, we take the bottom-up approach, and propose a simple method which requires only minimally more computation than the current state-of-the-art approaches for categorical semantic segmentation.

The bottom-up approach to instance segmentation typically requires, after semantic segmentation, only the additional step of grouping pixels into instances. This often involves some predicted representation or embedding of the pixels in such a way that the pixels can be clustered into instances based on some distance metric applied to the resulting representations. The embedding space can optionally be learned in itself by the network by training with a metric-learning approach, as in~\cite{metric, discriminative, associative, recurrent-embedding}, but the downsides of this are that the training procedure can often be unstable, and the clustering required after inference for pixel-grouping can be expensive, especially if the embedding space is high-dimensional.

Another option is to train in a traditionally supervised manner, and fix the embeddings to some inherent geometric feature(s) of the instances themselves. This is done for example by ~\cite{DCM} wherein the pixel embeddings correspond to their distance from the center-of-mass of their corresponding instances. More recently, Papandreou \etal~\cite{PersonLab} embed the pixels, specifically for human instance segmentation, in a geometric feature space which corresponds to the human pose of the instance.

In this work, we choose the embedding space of the pixels to correspond to the bounding box coordinates of the instance to which each pixel belongs. Although in theory this is not the ideal encoding, as it is possible for multiple object instances to have very heavy overlap and thus almost identical bounding boxes, it is sufficient to distinguish between instances in a vast majority of ``in-the-wild" scenes, and lends itself to a very efficient implementation. Additionally, this method is easy to implement on top of any existing architecture designed for semantic segmentation.

Inspired by the single shot approaches to object detection, we represent the task of each pixel location to predict the offsets of its instance's bounding box from a fixed prior ``anchor" bounding box centered at its own location. In fact, this method can be thought of as a ``dense" object detector, in which each pixel is allowed a proposal.

These predicted bounded-box embeddings can then be used to group the pixels into instances using an efficient algorithm based on the popular metric for the bounding box space, the Jaccard index, or ``intersection-over-union" (IoU).

Although this approach can in theory be used on any arbitrary number of object classes, as bounding box regressors can be class-agnostic~\cite{SSD, Retinanet}, we test our approach only on the person class as it allows for quicker training and easier evaluation, and person class results on the COCO dataset for several state-of-the-art methods have been reported in~\cite{PersonLab}.

\section{Related Work}

Approaches to instance segmentation can be divided into the top-down and bottom-up formulations. In top-down approaches, instance masks are obtained by refining bounding-box-level detections as in~\cite{FCIS, cascades, masklab}. The most successful of these has been the Mask-RCNN~\cite{MaskRCNN} system, which extends the Faster-RCNN object detector by predicting masks for each region proposal.

Many other recent approaches are of the bottom-up formulation, in which pixels are first classified and afterwards assigned to instances. Liang \etal~\cite{proposal-free} propose to learn pixel-wise regional features and class-wise instance numbers which can be used to cluster the classified pixels. Uhrig \etal~\cite{depth-layering} predict direction to instance centers for each pixel and then use template matching on the rectangular directional patterns.
Wu \etal~\cite{bridging} predict semantic masks as well as bounding boxes for each pixel, then group using a Hough-like voting scheme. Our approach is most similar to~\cite{bridging}, the main differences being the introduction of bounding box priors and the combination of the categorical classification and the bounding box regression into a single network with an off-the-shelf backbone architecture. Kirillov \etal~\cite{instancecut} propose Instancecut, which predicts instance boundaries for the purpose of separating pixels into instances. A metric learning approach to learn an embedding space that can be used for clustering pixels into instances has been proposed by~\cite{metric, discriminative, associative, recurrent-embedding}. Liu \etal propose the SGN, which employs a sequence of networks to perform inscreasingly complex gradual gouping into instances. Papandreou \etal~\cite{PersonLab} propose PersonLab, specifically for person instance segmentation, which predicts geometric features for each pixel based on its offsets from human pose keypoints which are also predicted by the network.

\section{Methods}

We develop a single-shot bottom-up approach to person instance segmentation.
Given an image, it consists of (1) classification of each pixel as person or background, and (2) grouping the person-classified pixels into person instances. Figure~\ref{fig:bbe-diagram} shows a diagram of the full method, which we will describe in further detail below.

\subsection{Semantic Person Segmentation}

We treat the segmentation portion of our approach in the standard fully-convolutional manner. We predict for each pixel location \bfxi\, the probability $p_S(\bfxi)$ that it belongs to a person instance. (For the COCO dataset, this includes all person annotations including crowd regions.)
At training time, we compute and backpropogate the loss as the average of the logistic loss over all pixel locations.

\subsection{Bounding Box Proposals}

\begin{figure}
	\begin{center}
		\includegraphics[width=0.9\linewidth]{anchor_box.png}
	\end{center}
	\caption{An anchor box and a ground truth box for a given peron pixel location.
		The purple dot shows the pixel location under consideration, and the broken-line box shows its associated anchor proposal box. The red box is the ground truth for that pixel. The network is trained to predict the offsets from the anchor to the ground truth at each pixel location.}
	\label{fig:anchorbox}
\end{figure}

For the purpose of predicting the person instance to which each pixel belongs, each pixel location \bfxi\, has associated with it a ``proposal" or ``anchor" bounding box which is centered at that pixel location and has a fixed canonical width \(a_w\) and height \(a_h\).
For each pixel, our network predicts the offsets \((d_x, d_y, d_w, d_h)\) from its anchor box to the bounding box of the instance to which it belongs. Figure~\ref{fig:anchorbox} shows an example of an anchor box and a ground truth box for a given pixel location.

We parametrize the bounding box offsets in the manner described in ~\cite{RCNN}. The offsets $d_x$ and $d_y$ describe scale-invariant translations from the center of the anchor box, while $d_w$ and $d_h$ describe log-space translations of the width and height of the anchor box.
During training, we use the instance mask annotations to construct per-pixel ground truth bounding box offsets. We then compute and backpropogate the $L_1$ loss between the predictions and the ground truth offsets only at locations corresponding to person pixels in the ground truth annotations.

\begin{figure*}
	\begin{center}
		\includegraphics[width=0.9\linewidth]{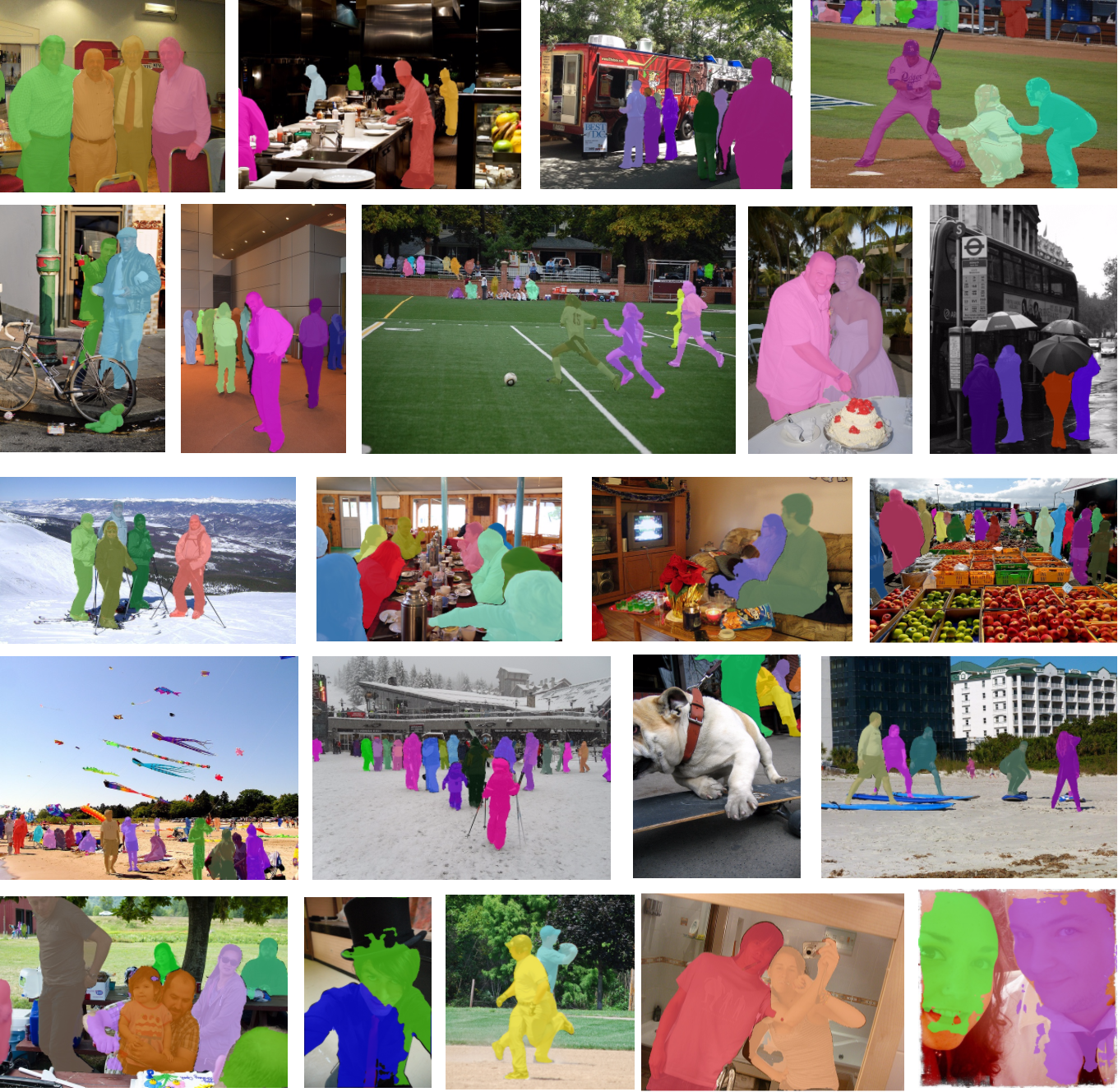}
	\end{center}
	\caption{Visualization of some results from the COCO \textit{val} split using our ResNet-50 model. We try to choose examples of scenes with multiple person instances which are overlapping. The colors of the instance overlays are randomly chosen. The last row shows some typical failure cases either in the categorical prediction or in the instance assignment.}
	\label{fig:qual_results}
\end{figure*}

\subsection{Instance Grouping}

Given the pixel-wise predicted bounding boxes, our instance grouping algorithm has two sequential steps: (1) Selecting the global bounding box detections which represent the person instances in the image, and (2) Assigning each person pixel to one of those global bounding boxes.

{\bf Selecting global boxes:} Given the semantic segmentation map predicted by our network,
we treat the probability value assigned to each location $p_S(\bfxi)$ as the confidence value associated with the bounding box predicted for that location.
We collect all predicted bounding boxes which correspond to local peaks in the semantic segmentation map and have a confidence score above a threshold $t_c$. (In our experiments, $t_c=0.6$.) Non-maximum suppression (with an IoU threshold of 0.4) is then applied to the collected bounding boxes to obtain the global bounding box detections $B_g$ for the given image. 

{\bf Assigning Pixels:} Let \(S_p\) be the set of all person pixels; that is, all pixels locations for which the semantic segmentation map provides a probability $p_S(\bfxi)\geq0.5$.
Each of these pixels needs to be assigned to one of the global instance bounding boxes $B_g$ from the previous stage.

For each pixel location \bfxi\, in \(S_p\) we take its corresponding predicted bounding box $\mathbf{b}_i$ and compute the IoU between $\mathbf{b}_i$ and each of the boxes in \(B_g\).
The pixel location is then assigned to the box in \(B_g\) with which it acheives the maximum IoU score which is above a threshold \(t_{iou}\).
If all of the boxes in \(B_g\) overlap with $\mathbf{b}_i$ with an IoU score of less than \(t_{iou}\), then the pixel location \bfxi is removed from \(S_p\); \ie it is assumed to be a false positive result of the semantic segmentation and is assigned to none of the instances.

Following the above two steps provides us with a set of boxes signifying person detections, with each box assigned a set of pixels which corresponds to the instance mask associated with that detection. For each detected instance, we compute the mean across its set of pixel locations of the predicted probabilities on the segmentation map, and assign that as the confidence score for the detection.

If there are $N_p$ person pixels and $M$ person instances in the image, our pixel assignment algorithm has complexity $\mathcal{O}(N_p \ast M)$, and runs extremely fast in practice (approximately 1-2 ms on an iPhone 7), introducing only neglible computation time on top of the forward pass of the network. Additionally, extention to multi-class applications is trivial, requiring only that the pixel grouping algorithm be performed per class.

\section{Experimental Evaluation}

\begin{table*}[t]
\caption{Performance on the person class of the COCO \textit{test-dev} split. Results from other methods have been obtained from~\cite{PersonLab}} 
\centering 
\begin{tabular}{l|c|lllllllllllll} 
\hline\hline \\ 
  & $AP$ & $AP^{50}$ & $AP^{75}$ & $AP^{S}$ & $AP^{M}$ & $AP^{L}$ & $AR^{1}$ & $AR^{10}$ & $AR^{100}$ & $AR^{S}$ & $AR^{M}$ & $AR^{L}$  \\ [0.5ex] 
\hline  \\
FCIS (ResNet-101) & 0.334 & 0.641 & 0.318 & 0.090 & 0.411 & 0.618 & 0.153 & 0.372 & 0.393 & 0.139 & 0.492 & 0.688 \\ 

PersonLab (ResNet-101) & 0.377 & 0.659 & 0.394 & 0.166 & 0.480 & 0.595 & 0.162 & 0.415 & 0.437 & 0.207 & 0.536 & 0.690 \\ 

BBE (Ours) (ResNet-50) & 0.368 & 0.628 & 0.374 & 0.125 & 0.440 & 0.622 & 0.162 & 0.390 & 0.406 & 0.174 & 0.498 & 0.673 \\ [1ex] 
\hline 
\end{tabular}
\label{table:results-test} 
\end{table*}

\begin{table*}[t]
\caption{Performance on the person class of the COCO \textit{val} split. Results from other methods have been obtained from~\cite{PersonLab}} 
\centering 
\begin{tabular}{l|c|lllllllllllll} 
\hline\hline \\ 
  & $AP$ & $AP^{50}$ & $AP^{75}$ & $AP^{S}$ & $AP^{M}$ & $AP^{L}$ & $AR^{1}$ & $AR^{10}$ & $AR^{100}$ & $AR^{S}$ & $AR^{M}$ & $AR^{L}$  \\ [0.5ex] 
\hline  \\
Mask-RCNN (ResNet-101) & 0.455 & 0.798 & 0.472 & 0.239 & 0.511 & 0.611 & 0.169 & 0.477 & 0.530 & 0.350 & 0.596 & 0.721 \\ 

PersonLab (ResNet-101) & 0.382 & 0.661 & 0.397 & 0.164 & 0.476 & 0.592 & 0.162 & 0.416 & 0.439 & 0.204 & 0.532 & 0.681 \\ 

BBE (Ours) (ResNet-50) & 0.374 & 0.644 & 0.386 & 0.127 & 0.449 & 0.616 & 0.160 & 0.399 & 0.418 & 0.180 & 0.511 & 0.665 \\ [1ex] 
\hline 
\end{tabular}
\label{table:results-val} 
\end{table*}

\subsection{Dataset}

We use the COCO dataset for training and evaluation.
We train only with the images from the \textit{train} set containing person annotations (64,115 images).
We report our results on the person class of the instance segmentation task in COCO for the \textit{val} and \textit{test-dev} splits (5,000 and 20,288 images respectively).

\subsection{Model Architecture}

We employ the ResNet-50~\cite{Resnet} base network, which we choose for its balance of feature richness and memory consumption, and attach to it the Atrous Spatial Pyramid Pooling module and the DeepLabv3+ decoder layers described in ~\cite{Deeplabv3, Deeplabv3+}. The only divergence from DeepLabv3+ is that in addition to the final 1x1 convolutional layer with one filter per class (in our case, there is only one class) on top of the decoder's output feature maps, we have an additional 1x1 convolutional layer with four filters to predict the dense bounding box offsets.

We train with output stride 16 and test with output stride 8, using atrous convolution in the later layers to retain receptive field, as described in~\cite{Deeplabv3}.

\subsection{Prior Boxes}

We use the annotations for the COCO \textit{val} split to empirically estimate a good prior bounding box size for the average person instance. We thus set the scale of the prior box at all pixel locations to be $96^{2}$ pixels with an aspect ratio of 1.5 (person instances on average have greather height than width). The prior box for each pixel location is centered at that location itself, as shown in Figure~\ref{fig:anchorbox}.

\subsection{Training Procedure}

We use the Tensorflow~\cite{tensorflow} and Keras~\cite{keras} software packages to implement our method.

During training, we apply data augmentation by randomly resizing the input image and associated ground truth masks by a scale factor between 0.5 and 1.5, rotating up to 15 degrees, and translating vertically and horizontally by up to \(\frac{1}{10}\) of the final crop size, which we set at 801x801 pixels.

We initialize the weights of the ResNet-50 base network with those pre-trained on the Imagenet classification task, and distribute a batch size of 12 images across four NVIDIA GTX 1080 Ti GPUs. We then train with the Adam optimizer ~\cite{Adam} with a base learning rate of \(10e^{-4}\) for 185K iterations. Then, we lower the base learning rate to $10e^{-5}$ and train for another 80K iterations.

\begin{table*}[t]
\caption{MobileNetV2 Results on the COCO \textit{val} split person class} 
\centering 
\begin{tabular}{c|c|lllllllllllll}
\hline\hline \\ 
  OS & ASPP & $AP$ & $AP^{50}$ & $AP^{75}$ & $AP^{S}$ & $AP^{M}$ & $AP^{L}$ & $AR^{1}$ & $AR^{10}$ & $AR^{100}$ & $AR^{S}$ & $AR^{M}$ & $AR^{L}$  \\ [0.5ex] 
\hline  \\
16 && 0.246 & 0.490 & 0.228 & 0.047 & 0.269 & 0.492 & 0.138 & 0.301 & 0.311 & 0.085 & 0.377 & 0.576 \\

8 && 0.260 & 0.507 & 0.242 & 0.056 & 0.292 & 0.501 & 0.141 & 0.316 & 0.327 & 0.101 & 0.399 & 0.583 \\

16 &\checkmark & 0.280 & 0.531 & 0.270 & 0.054 & 0.311 & 0.548 & 0.143 & 0.316 & 0.327 & 0.093 & 0.394 & 0.604 \\ 

8 &\checkmark & 0.296 & 0.553 & 0.285 & 0.068 & 0.331 & 0.557 & 0.145 & 0.331 & 0.343 & 0.115 & 0.414 & 0.604 \\ 
[1ex] 
\hline
\end{tabular}
\label{table:mobilenet_results}
\end{table*}

\subsection{Results}

Tables~\ref{table:results-test} and~\ref{table:results-val} show our results on the person class of the \textit{test-dev} and \textit{val} splits of the COCO dataset. We use single scale inference, resizing each image so that its larger dimension has size 961. We limit to a maximum of 20 person proposals per image as done in~\cite{PersonLab}.

For the \textit{test-dev} split, we compare against the top-down FCIS method~\cite{FCIS}, winner of the 2016 COCO detection challenge, which our method outperforms by $3.4\%$ on the person class. For the \textit{val} split, we compare against Mask-RCNN~\cite{MaskRCNN}, the most successful of the top-down approaches, which outperforms our method by a large margin of $8.2\%$. We note, though, that our method outperforms all other methods including Mask-RCNN at the mAP score for large instances.

On both the \textit{test-dev} and \textit{val} splits, we compare against the bottom-up PersonLab~\cite{PersonLab} method (evaluated at single-scale inference), which performs joint human pose estimation and human instance segmentation. Our method is comparable in accuracy to PersonLab, which outperforms our method by less than $1\%$ on both dataset splits. We point out that our comparison is not completely pure. On the one hand, the PersonLab evaluation involves a deeper ResNet-101 backbone (compared to our ResNet-50) and a test resolution of 1401 pixels (compared to our 961); and on the other hand our method incoporates the Deeplabv3+ decoder module from~\cite{Deeplabv3+} whereas PersonLab directly upsamples the outputs from an output stride of 8.

We stress here that the main advantage of our approach is its efficiency. It does not require a two-stage system like FCIS or Mask-RCNN, nor does it require the keypoint predictions and offsets (as well as the bilinear offset-refinement stage) of PersonLab.

\subsection{MobileNetV2 Results}
To show that our approach can be compatible with mobile devices with restrictive computational power, we also train the same model with a MobileNetV2~\cite{mobilenetv2} backbone, which is specifically designed for performance on mobile devices. We follow the suggestions in Section 6.3 of~\cite{mobilenetv2} in regard to architectural decisions and experiment with and without the full Atrous Spatial Pyramid Pooling module.

We perform a similar training procedure to the above, and report our results for person instance segmentation using output strides of 8 and 16 on the \textit{val} split of the COCO dataset in Table~\ref{table:mobilenet_results}.

\subsection{Qualitative Results}

Figure~\ref{fig:qual_results} shows examples of the results of our method on the person class from the \textit{val} split of the the COCO dataset using single-scale inference.

\section{Conclusion}
We've presented a single-shot method for object instance segmentation, and showed its effectiveness at the task of person instance segmentation. Our approach is virtually as efficient as categorical semantic segmentation and opens up the possibility of real-time applications on embedded and mobile devices.

{\small
\bibliographystyle{ieee}
\bibliography{bbe_segmentation}
}

\end{document}